\documentclass{article}

\usepackage[final]{neurips_2024}

\usepackage[utf8]{inputenc}
\usepackage[T1]{fontenc}
\usepackage{url}
\usepackage{booktabs}
\usepackage{amsfonts}
\usepackage{amsmath}
\usepackage{amssymb}
\usepackage{nicefrac}
\usepackage{microtype}
\usepackage[dvipsnames]{xcolor}
\usepackage{graphicx}
\usepackage{subcaption}
\usepackage{wrapfig}
\usepackage{algorithm}
\usepackage{algorithmic}
\usepackage{multirow}
\usepackage{makecell}
\usepackage{colortbl}
\usepackage{enumitem}
\usepackage{xspace}
\usepackage{pifont}
\usepackage{tcolorbox}
\tcbuselibrary{breakable, skins}

\definecolor{mj1blue}{RGB}{41, 128, 185}
\definecolor{mj1green}{RGB}{39, 174, 96}
\definecolor{mj1red}{RGB}{192, 57, 43}
\definecolor{mj1purple}{RGB}{142, 68, 173}
\definecolor{codegreen}{rgb}{0,0.6,0}
\definecolor{codegray}{rgb}{0.5,0.5,0.5}
\definecolor{codepurple}{rgb}{0.58,0,0.82}
\definecolor{backcolour}{rgb}{0.97,0.97,0.97}
\usepackage{hyperref}
\hypersetup{
  colorlinks=true,
  urlcolor=mj1red
}

\usepackage{listings}
\lstdefinestyle{xmlstyle}{
    backgroundcolor=\color{backcolour},
    commentstyle=\color{codegreen},
    keywordstyle=\color{mj1blue}\bfseries,
    numberstyle=\tiny\color{codegray},
    stringstyle=\color{codepurple},
    basicstyle=\ttfamily\footnotesize,
    breakatwhitespace=false,
    breaklines=true,
    captionpos=b,
    keepspaces=true,
    numbers=left,
    numbersep=5pt,
    showspaces=false,
    showstringspaces=false,
    showtabs=false,
    tabsize=2,
    frame=single,
    framerule=0.5pt,
    rulecolor=\color{gray!30},
}

\usepackage{xcolor}
\definecolor{mjcolor}{RGB}{180,20,30}
\newcommand{\mj}{\textbf{\textcolor{mjcolor}{MJ$_1$}}\xspace}

\usepackage{amsthm}

\usepackage{tikz}
\usetikzlibrary{shapes.geometric, arrows, positioning, fit, backgrounds, calc}

\title{
\begin{center}
\raisebox{-0.7em}{\includegraphics[height=2em]{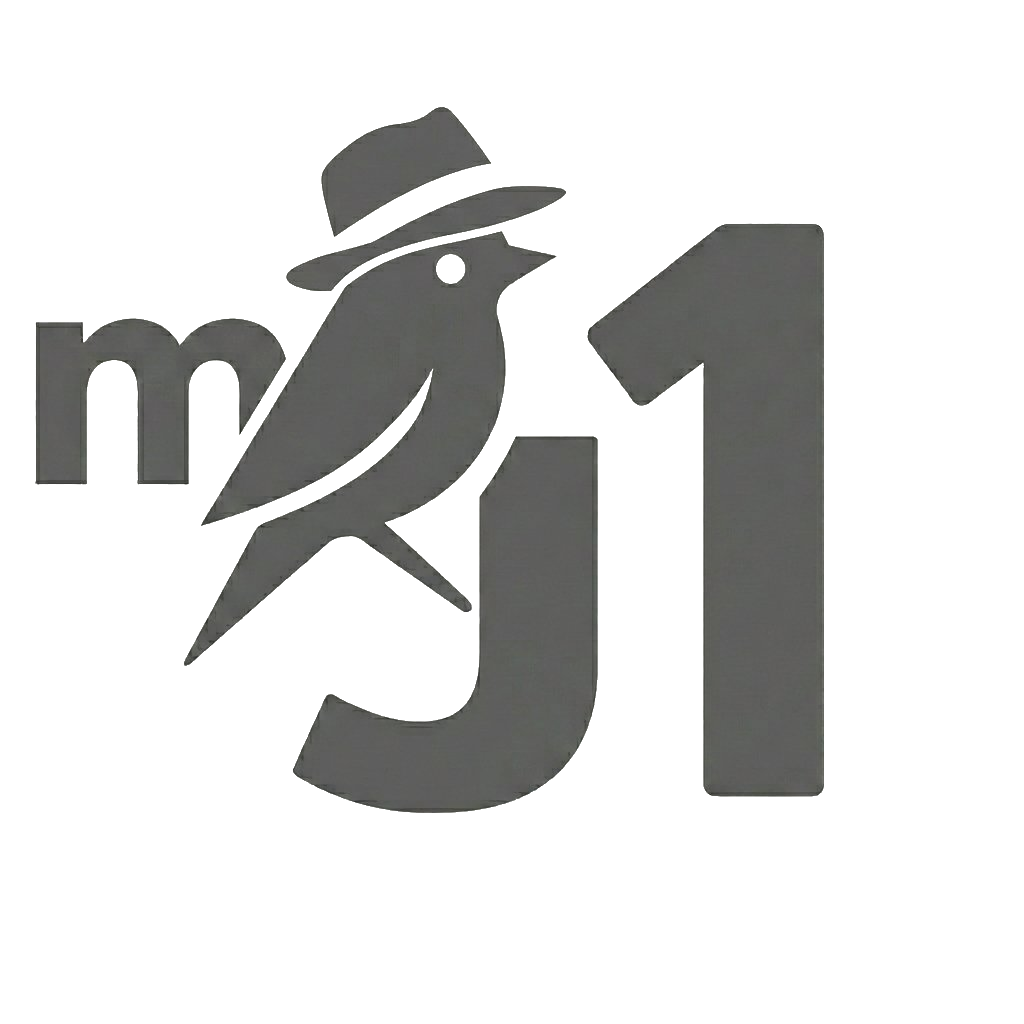}}
\hspace{-0.6em}
\textbf{: Multimodal Judgment via \textit{\textcolor{gray}{Grounded Verification}}}
\end{center}
\vspace{-0.5em}
}

\newcommand{\haizebird}{\raisebox{-4.5pt}{\includegraphics[height=16pt]{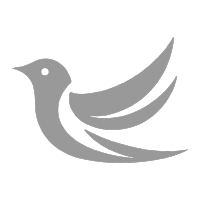}}}

\author{
  Bhavesh Kumar, Dylan Feng, Leonard Tang \\
  \haizebird\, Haize Labs \\
  \texttt{\{bhavesh, dylan, leonard\}@haizelabs.com}
}

\begin{document}

\maketitle

\begin{abstract}

Multimodal judges struggle to ground decisions in visual evidence. We present \mj, a reinforcement-learning--trained multimodal judge that enforces visual grounding through a structured \textit{grounded verification chain} (observations $\rightarrow$ claims $\rightarrow$ verification $\rightarrow$ evaluation $\rightarrow$ scoring) and a counterfactual consistency reward that penalizes position bias. Even without training, our mechanism improves base-model accuracy on MMRB2 by +3.8 points on Image Editing and +1.7 on Multimodal Reasoning. After training, \mj, with only 3B active parameters, achieves 77.0\% accuracy on MMRB2 and surpasses orders-of-magnitude larger models like Gemini-3-Pro. These results show that grounded verification and consistency-based training substantially improve multimodal judgment without increasing model scale. A demo of \mj can be found at \href{https://mj1.haizelabs.com/}{mj1.haizelabs.com}.

\end{abstract}

\section{Introduction}
\label{sec:intro}

The ability to measure the extent to which generated images satisfy a user's intent is core to how we align, evaluate, and improve vision-language models. It underlies reward modeling for RLHF \citep{ouyang2022training}, automated benchmark evaluation \citep{zheng2023judging}, and data quality filtering at scale. However, despite its criticality, multimodal judge performance lags behind text judges. On Multimodal RewardBench~2 (MMRB2), the current most comprehensive multimodal judgement benchmark, frontier models like Gemini-3-Pro and GPT-5 achieve only 70--76\% accuracy, and the best open-source models saturate near 64\% \citep{hu2025mmrb2}. Multimodal RewardBench and VL-RewardBench offer similar results: both frontier and fine-tuned open-source models underperform on tasks requiring sustained visual reasoning \citep{yasunaga2025mmrb1, li2024vlrewardbench}. These results suggest the bottleneck is not model scale but rather a mechanical failure in how VLMs process and reason about visual evidence.

Multiple prior works investigate this failure. FastV demonstrates that visual tokens receive vanishingly small attention weights in deeper transformer layers and can be pruned by 50\% after layer~2 with negligible performance loss. Visual information largely stops propagating well before the model's final layers \citep{chen2024fastv}. SparseVLM finds that visual tokens carry sparser information density than text tokens and that text tokens effectively gate which visual tokens receive attention at all \citep{zhang2024sparsevlm}. LLaVA-Mini extends this to an extreme, compressing 576 visual tokens to a single token while matching full performance by pre-fusing visual content into text representations in the earliest layers \citep{zhang2025llavamini}. Concurrently, \citet{fu2025hidden} show that VLMs perform dramatically worse than their own visual encoders on vision-centric tasks because the VLM over-attends language priors. \citet{han2025selfconsistency} demonstrate that the same model that hallucinates objects in long detailed captions will correctly deny those hallucinations when directly questioned, suggesting that visual knowledge is encoded but becomes inaccessible as generation extends. These failures are amplified in multimodal judgment, which requires simultaneous processing of multiple images and extended reasoning to determine human preference.

Recent work on training thinking judges via reinforcement learning has shown that RL-trained judges with explicit chain-of-thought dramatically outperform SFT-based approaches, particularly on reasoning-intensive evaluation tasks. J1 \citep{whitehouse2025j1} introduces a unified RL framework for training text-domain judges with verifiable rewards, achieving state-of-the-art performance across multiple text benchmarks. JudgeLRM \citep{chen2025judgelrm} independently confirms that SFT benefits negatively correlate with reasoning difficulty, and that GRPO-trained judges at 3B--7B surpass GPT-4 and DeepSeek-R1. EvalPlanner \citep{saha2025evalplanner} separates evaluation into explicit planning and execution phases, achieving strong results with minimal synthetic data. However, all of these approaches operate exclusively in the text domain. Extending RL-trained thinking judges to multimodal judgment introduces a qualitatively different challenge: the judge must not only reason well but must do so while respecting visual evidence across multiple images, where attention decay is most severe.

We address this with \mj, which makes two contributions:
\begin{enumerate}[leftmargin=*, itemsep=2pt]
\item A \emph{grounded verification chain} that decomposes multimodal judgment into a structured sequence of stages. The model first extracts visual observations from each image when text context is minimal and visual attention is highest; extracts claims from each response; verifies claims against observations; evaluates claims against task-specific criteria; and finally produces a final score. We show that this structured prompting alone improves accuracy by +3.8 points on MMRB2 Image Editing and +1.7 points on Multimodal Reasoning over open-ended prompting (Section~\ref{sec:empirical}).

\item A \emph{counterfactual consistency reward} for training position-invariant multimodal judges. Extending J1's insight \citep{whitehouse2025j1} that consistency-based rewards mitigate positional bias in text judges, we enforce answer invariance under swapped image inputs. Prior to this reward, the model selected Response A roughly twice as often as Response B within each training batch despite balanced ground-truth labels. The consistency reward largely eliminated this to near parity.
\end{enumerate}

We train Qwen3-VL-30B-A3B \citep{qwen2025qwen3vl} via SFT on distilled reasoning traces followed by GRPO \citep{shao2024deepseekmath} with a three-component reward covering format compliance, correctness, and counterfactual consistency. \mj achieves state-of-the-art performance on MMRB2, surpassing Gemini-3-Pro with orders of magnitude less parameters.

\section{Methodology}
\label{sec:method}

A (preference) multimodal judge receives a prompt $p$, and two candidate responses $R_A$ and $R_B$. $p$, $R_A$, and $R_B$ can each contain both text and images. The judge's task is to determine which response better fulfills the prompt. Standard autoregressive judgement produces a final score at the end extended text generation where attention to visual tokens has substantially decayed \citep{chen2024fastv, huang2024opera}. These scores are often not \textit{grounded} in the input images.

\subsection{Grounded Verification Chain}
\label{sec:chain}

To combat visual attention degradation, \mj generates answers in the following sequence:
\begin{equation}
(p, R_A, R_B) \xrightarrow{g_O} O \xrightarrow{g_C} C \xrightarrow{g_V} V \xrightarrow{g_E} E \xrightarrow{g_s} s
\label{eq:chain}
\end{equation}

The five stages proceed as follows. First, in the \emph{visual observation} stage ($O$), the model describes the visual content of images in $p$, $R_A$, and $R_B$. Second, in \emph{claim extraction} ($C$), the model decomposes $R_A$ and $R_B$ into claims. Third, in \emph{consistency verification} ($V$), each claim is verified against the observations from $O$. This produces a binary signal: $1$ for claim-observation consistency and $0$ otherwise. This forces the reasoning to attend to the initial visual evidence. Fourth, in \emph{criteria evaluation} ($E$), the model evaluates both responses against task-specific criteria (see Appendix \ref{app:prompts}). Finally, in \emph{scoring} ($s$), the model produces integer scores $\{s_A, s_B\}$ where $s_A, s_B \in [1, 10]$ and $s_A \neq s_B$.

\begin{figure}[http]
    \centering
    \resizebox{0.95\textwidth}{!}{%
    \begin{tikzpicture}[
        node distance=0.6cm,
        box/.style={rectangle, draw, rounded corners, minimum width=2.2cm, minimum height=0.8cm, align=center, font=\scriptsize},
        arrow/.style={->, >=stealth, thick}
    ]
        \node[box, fill=mj1blue!20] (img) {Prompt $p$ + \\ Responses $R_A, R_B$};
        \node[box, fill=mj1green!20, right=0.6cm of img] (obs) {\textbf{Observations}\\$O_p, O_A, O_B$};
        \node[box, fill=mj1purple!20, right=0.6cm of obs] (claims) {\textbf{Claims}\\$C_A, C_B$};
        \node[box, fill=orange!20, right=0.6cm of claims] (verify) {\textbf{Verification}\\$V_A, V_B$};
        \node[box, fill=mj1blue!12, right=0.6cm of verify] (eval) {\textbf{Evaluation}\\per criterion};
        \node[box, fill=mj1red!20, right=0.6cm of eval] (score) {\textbf{Scores}\\$\boxed{s_A, s_B}$};

        \draw[arrow] (img) -- (obs);
        \draw[arrow] (obs) -- (claims);
        \draw[arrow] (claims) -- (verify);
        \draw[arrow] (verify) -- (eval);
        \draw[arrow] (eval) -- (score);

        \draw[arrow, dashed, mj1green!70!black, thick] (verify.south) to[out=-90, in=-90, looseness=0.4] node[midway, below, font=\tiny] {Visual Grounding} (obs.south);
    \end{tikzpicture}%
    }
    \caption{\textbf{\mj grounded verification chain.} Judgement scores are generated based on verifying response claims against visual observations. Explicit visual grounding of the reasoning chain mitigates visual attention degradation.}
    \label{fig:architecture}
\end{figure}

The combination of early visual observation and consistency verification between reasoning and visual observations is the key advantage that enables \mj's state-of-the-art performance.

\subsection{Training Pipeline}
\label{sec:training}
\label{sec:consistency}

Training proceeds in two phases (Figure~\ref{fig:training}). A cold-start SFT phase is followed by GRPO \citep{shao2024deepseekmath} with the following composite reward, where $J$ is \mj's prediction and $y^*$ is the ground-truth label:

\begin{equation}
R(J) = R_{\text{format}}(J) + R_{\text{correct}}(J, y^*) + R_{\text{cons}}(J, J', y^*)
\label{eq:reward}
\end{equation}

\begin{figure}[http]
    \centering
    \resizebox{0.65\textwidth}{!}{%
    \begin{tikzpicture}[
        node distance=1.2cm,
        phase/.style={rectangle, draw, rounded corners=4pt, minimum width=3.2cm, minimum height=1cm, align=center, font=\footnotesize},
        reward/.style={rectangle, draw, rounded corners=3pt, minimum width=2.4cm, minimum height=0.55cm, align=center, font=\tiny},
        arrow/.style={->, >=stealth, thick}
    ]
        \node[phase, fill=mj1blue!15] (sft) {\textbf{Cold-Start SFT}\\{\scriptsize 10k Distilled traces}};
        \node[phase, fill=mj1green!20, right=2cm of sft] (grpo) {\textbf{GRPO}\\{\scriptsize 1 epoch}};
        \node[reward, fill=mj1green!10, below=0.9cm of grpo, xshift=-3cm] (r1) {$R_{\text{format}} \in [0, 0.2]$\\XML tag compliance};
        \node[reward, fill=mj1green!10, below=0.9cm of grpo] (r2) {$R_{\text{correct}} \in \{0, 1\}$\\Preference accuracy};
        \node[reward, fill=mj1green!10, below=0.9cm of grpo, xshift=3cm] (r3) {$R_{\text{cons}} \in \{0, 1\}$\\Counterfactual flip};

        \draw[arrow] (sft) -- (grpo);
        \draw[arrow, mj1green!60] (r1.north) -- ([xshift=-1.2cm]grpo.south);
        \draw[arrow, mj1green!60] (r2.north) -- (grpo.south);
        \draw[arrow, mj1green!60] (r3.north) -- ([xshift=1.2cm]grpo.south);
    \end{tikzpicture}%
    }
    \caption{\textbf{Two-phase training pipeline.} Cold-start SFT on distilled reasoning traces establishes format and basic judgment capability. GRPO then optimizes a composite reward that incentivizes both correctness and position invariance.}
    \label{fig:training}
\end{figure}

The format reward $R_{\text{format}} \in [0, 0.2]$ validates XML structure, assigning $\tfrac{0.2}{11}$ per correctly formed tag across 11 required tags (5 standalone sections plus 2 parent sections each with 2 nested sub-tags). If score parsing fails entirely (no valid $\boxed{s_A, s_B}$ extracted), the total reward is set to zero regardless of format compliance, ensuring the model cannot earn reward without producing a parseable judgment. The correctness reward $R_{\text{correct}} \in \{0, 1\}$ indicates whether $sign(s_A - s_B)$ matches the ground-truth label $y^*$, with no ties allowed ($s_A \neq s_B$).

The consistency reward $R_{\text{cons}} \in \{0, 1\}$ mitigates positional bias. J1 \citep{whitehouse2025j1} introduced consistency-based rewards for mitigating positional bias in text judges, granting a reward when the model produces the correct verdict under both orderings of a response pair. We extend this idea to the multimodal setting under our grounding mechanism.

During GRPO, we swap the inputs to \mj and also swap all references of $R_A$ to $R_B$ in the response understanding, claim, and verification sections. The original evaluation and scores are discarded. The model resumes reasoning from the truncation point with temperature 0, regenerating only the evaluation and scoring stages rather than the full reasoning chain. We check whether the preference correctly inverts: $R_{\text{cons}} = 1$ if it does, and $R_{\text{cons}} = 0$ otherwise. 

For each training point, we generate a group of 32 completions at temperature 0.7. Each completion is copied to calculate the positional consistency reward, yielding a total group size of 64. Advantages are then computed as group-relative deviations from the mean reward, pooling both original and flipped generations:
\begin{equation}
\hat{A}_i = R_i - \frac{1}{|\mathcal{G}|} \sum_{j \in \mathcal{G}} R_j
\end{equation}
where $\mathcal{G}$ includes both the original 32 completions and their corresponding flipped continuations. Forward-backward passes are filtered to sequences with $|\hat{A}_i| > 0.01$, avoiding near-zero-gradient updates. Samples for which all original completions have zero accuracy are skipped entirely, preventing reward hacking on format-only signal. We use LoRA with rank~64 and a cosine learning rate schedule ($5 \times 10^{-5} \to 1 \times 10^{-7}$, 10\% warmup).

\section{Analysis}
\label{sec:theory}

This section provides a formal analysis of why the \mj architecture incentivizes visual grounding. We first characterize visual grounding failure (Section~\ref{sec:problem}), then analyze how \mj's structure blocks the computational shortcuts that enable it (Section~\ref{sec:structural}), and finally validate these claims empirically (Section~\ref{sec:empirical}).

\subsection{Visual Grounding Failure in Judgment}
\label{sec:problem}

Consider the multimodal judgment task: given a prompt $p$ and two candidate responses $R_A, R_B$, a judge produces scores $s_A, s_B \in [1, 10]$. Let $y^* \in \{A, B\}$ denote the ground-truth preference. The judge succeeds when $\text{sign}(s_A - s_B)$ matches $y^*$.

We say a judge exhibits \emph{visual grounding failure} when it achieves above-chance accuracy while ignoring image contents $\mathcal{I}$ of $p, R_A, R_B$. Formally, let $\pi_\theta(s \mid p, R_A, R_B)$ denote the judge's scoring distribution. Grounding failure occurs when
\begin{equation}
\pi_\theta(s \mid p, R_A, R_B) \approx \pi_\theta(s \mid p, R_A, R_B \setminus \mathcal{I})
\label{eq:grounding_failure}
\end{equation}
meaning that the score distribution is approximately invariant to the image. This pathology arises when text-only features such as response fluency, length, or formatting correlate with quality in the training distribution, enabling the model to exploit shortcut features \citep{geirhos2020shortcut}. The attention decay mechanism provides a physical basis whereby, as generation proceeds, attention to image tokens decreases monotonically \citep{huang2024opera, jiang2025devils}, and scoring tokens appear at the end of extended outputs when image attention is minimal.

\begin{figure}[http]
    \centering
    \begin{tikzpicture}[
        node distance=0.6cm and 1.0cm,
        box/.style={rectangle, draw, rounded corners=3pt, minimum width=1.6cm, minimum height=0.6cm, align=center, font=\footnotesize},
        inputbox/.style={box, fill=mj1blue!15},
        obsbox/.style={box, fill=mj1blue!25},
        claimbox/.style={box, fill=mj1purple!18},
        verbox/.style={box, fill=mj1green!25},
        scorebox/.style={box, fill=mj1green!15},
        arrow/.style={->, >=stealth, thick},
        shortcut/.style={->, >=stealth, line width=1.3pt, dashed, red!65},
    ]
        \node[font=\small\bfseries] at (-2.2, 0) {(a) Standard};
        \node[inputbox] (img1) at (0.5, 0.55) {$p, R_A, R_B$};
        \node[inputbox] (text1) at (0.5, -0.55) {$p, R_A, R_B \setminus \mathcal{I}$};
        \node[scorebox] (score1) at (5.5, 0) {Scores $s$};
        \draw[arrow] (img1.east) -- ([yshift=0.12cm]score1.west);
        \draw[shortcut] (text1.east) -- ([yshift=-0.12cm]score1.west);
        \node[font=\scriptsize, red!65] at (3.0, -0.85) {shortcut};

        \node[font=\small\bfseries] at (-2.2, -2.8) {(b) MJ1};
        \node[inputbox] (img2) at (0.5, -2.25) {Images $\mathcal{I}$};
        \node[inputbox] (text2) at (0.5, -3.35) {$p, R_A, R_B \setminus \mathcal{I}$};
        \node[obsbox] (obs) at (3.0, -2.8) {Obs.\ $O$};
        \node[claimbox] (claims) at (5.2, -2.8) {Claims $C$};
        \node[verbox] (ver) at (7.4, -2.8) {Verif.\ $V$};
        \node[scorebox] (score2) at (9.6, -2.8) {Scores $s$};

        \draw[arrow] (img2.east) -- (obs.north west);
        \draw[arrow] (text2.east) -- (obs.south west);
        \draw[arrow] (obs.east) -- (claims.west);
        \draw[arrow] (claims.east) -- (ver.west);
        \draw[arrow] (ver.east) -- (score2.west);

        \draw[arrow, dashed, mj1green!70!black, thick]
            (ver.south) to[out=-55, in=-125, looseness=0.35]
            node[midway, below, font=\tiny, yshift=-0.06cm] {Visual Grounding}
            (obs.south);

        \draw[<->, mj1green!70, thick, line width=1.1pt]
            ([yshift=-0.06cm]ver.south east)
            to[out=-45, in=-135, looseness=1.0]
            node[below, font=\scriptsize, yshift=-0.06cm] {$R_{\text{cons}}$}
            ([yshift=-0.06cm]score2.south west);
    \end{tikzpicture}
    \caption{\textbf{Computational structure comparison.} (a) Standard judgment permits a shortcut path (dashed red) where scores depend minimally on images. (b) \mj forces computation through observations $O$, claim extraction $C$, and verification $V$. The dashed arrow indicates the forced back-reference from verification to observations. The consistency reward $R_{\text{cons}}$ couples verification to scores, requiring coherent image-grounded reasoning.}
    \label{fig:dag}
\end{figure}

\subsection{Structural Resistance to Shortcuts}
\label{sec:structural}

The grounded verification chain and counterfactual consistency reward interact to make visual grounding the path of least resistance. In \mj, scores $s$ are computed by the function composition $g_s \circ g_E \circ g_V \circ g_C \circ g_O$. The verification stage $g_V$ evaluates consistency between claims $C$ (from responses) and observations $O$ (from images). A shortcut policy that generates observations independent of image content produces $O$ that are generic or hallucinated. When evaluating claim-observation consistency, such observations will be correct on easy samples where text alone suffices,but uncorrelated with ground truth on hard samples where image evidence is necessary. This creates a natural curriculum in which shortcut policies perform adequately on easy samples but are penalized via $R_{\text{correct}}$ on hard ones, generating a gradient signal toward grounded observation extraction.

The flip mechanism directly tests whether scores track content or position. Consider a biased policy $\pi_{\text{pos}}$ that tends to assign higher scores to whichever response appears first. When we swap A$\leftrightarrow$B in both the input and the reasoning, $\pi_{\text{pos}}$ will still prefer the first-position response, which now contains different content. The flip check detects this as $R_{\text{cons}} = 0$. The only way to consistently achieve $R_{\text{cons}} = 1$ is to produce evaluations that depend on response content rather than response position.

The two mechanisms are synergistic. The chain structure ensures that the model's reasoning contains explicit A/B-separated observations, claims, and sccores. The counterfactual flip exploits this structure by swapping the A/B assignments and checking whether the judgment tracks the swap. A model that generates visually grounded observations will produce scores that correctly identify which response's claims align with visual evidence. When the swap occurs, these verdicts correctly invert, supporting the correct flipped judgment. A model that generates ungrounded observations will produce arbitrary verdicts that do not systematically invert, causing flip failure. Thus $R_{\text{cons}}$ preferentially reinforces trajectories where the chain was genuinely grounded.

Under GRPO, we optimize $J(\theta) = \mathbb{E}_{x \sim \mathcal{D}} \mathbb{E}_{J \sim \pi_\theta(\cdot \mid x)}[R(J)]$ with policy gradient:
\begin{equation}
\nabla_\theta J = \mathbb{E} \left[ \nabla_\theta \log \pi_\theta(O, C, V, E, s \mid x) \cdot R(J) \right]
\end{equation}
For autoregressive generation, $\log \pi_\theta(J \mid x) = \sum_t \log \pi_\theta(J_t \mid J_{<t}, x)$, distributing gradient across all generation steps. When $R_{\text{cons}}$ is high, the entire trajectory including early observation extraction is reinforced. When $R_{\text{cons}}$ is low, the trajectory is penalized. This creates gradient coupling between late-stage scores and early-stage observations, mediated by the verification chain.

We note two important caveats. First, our argument establishes that visual grounding is a sufficient condition for high reward, not that it is strictly necessary. In principle, a model could achieve high $R_{\text{cons}}$ through elaborate ``consistency theater,'' generating internally coherent but image-independent outputs that happen to flip correctly. We argue that this is difficult, since the model must commit to observations before generating claims and verdicts in the autoregressive order, analogous to the difficulty of faking chain-of-thought reasoning \citep{lanham2023measuring}. However, we do not formally rule this possibility out. 

Second, the effectiveness of $R_{\text{cons}}$ depends on the training distribution containing samples where image evidence is necessary for correct judgment. On purely text-discriminable samples, the flip reward provides no additional grounding signal beyond what $R_{\text{correct}}$ already supplies.

\subsection{Empirical Validation}
\label{sec:empirical}

We validate our analysis with two experiments on untrained base models, isolating the effects of the grounding chain and the consistency mechanism independently of \mj training.

\textbf{Grounding chain improves judgment without training.} We compare two prompting strategies applied to the untrained Qwen3-VL-30B-A3B base model on a subset of 500 samples from MMRB2's Image Editing and Multimodal Reasoning subtasks. The first uses open-ended reasoning where the model is instructed to compare the responses and select the better one without structural guidance. The second uses the \mj grounding prompt described in Section~\ref{sec:chain}, which instructs the model to extract observations, claims, and verification before scoring.

\begin{table}[t]
\centering
\caption{\textbf{Effect of grounded verification prompting on untrained base model.} Accuracy (\%) on 500 samples each from MMRB2 Image Editing and Multimodal Reasoning. Structured grounding improves accuracy without any training.}
\label{tab:grounding_ablation}
\begin{tabular}{@{}lcc@{}}
\toprule
\textbf{Prompting Strategy} & \textbf{Image Editing} & \textbf{Multimodal Reasoning} \\
\midrule
Open-ended reasoning & 62.4 \phantom{\textcolor{ForestGreen}{(+3.8)}} & 53.4 \phantom{\textcolor{ForestGreen}{(+1.7)}} \\
\mj grounded verification & 66.2 {\textcolor{ForestGreen}{(+3.8)}} & 55.1 {\textcolor{ForestGreen}{(+1.7)}} \\
\bottomrule
\end{tabular}
\end{table}

Table~\ref{tab:grounding_ablation} shows that structured grounding improves accuracy by +3.8 points on Image Editing and +1.7 points on Multimodal Reasoning with zero training. This is consistent with our hypothesis that front-loading visual observation extraction when attention to image tokens is highest preserves visual information that would otherwise be lost during extended open-ended generation. The larger gain on Image Editing likely reflects that editing tasks require fine-grained visual comparison (e.g., detecting whether a specific edit was applied correctly), where explicit observation extraction is particularly valuable.

\textbf{Consistency reward correlates with visual grounding.} We prompt Qwen3-VL-30B-A3B with the \mj structured format (no training) and evaluate the same MMRB2 subset under three image conditions: (1) \emph{real images}, where each sample receives its correct corresponding image; (2) \emph{shuffled images}, where images are randomly permuted such that each sample receives an image from a different sample; and (3) \emph{blank image}, where the model receives a blank grey square as input. For each condition we measure both $R_{\text{cons}}$ and $R_{\text{correct}}$.

\begin{figure}[http]
    \centering
    \begin{subfigure}[b]{0.48\textwidth}
        \centering
        \includegraphics[width=\textwidth]{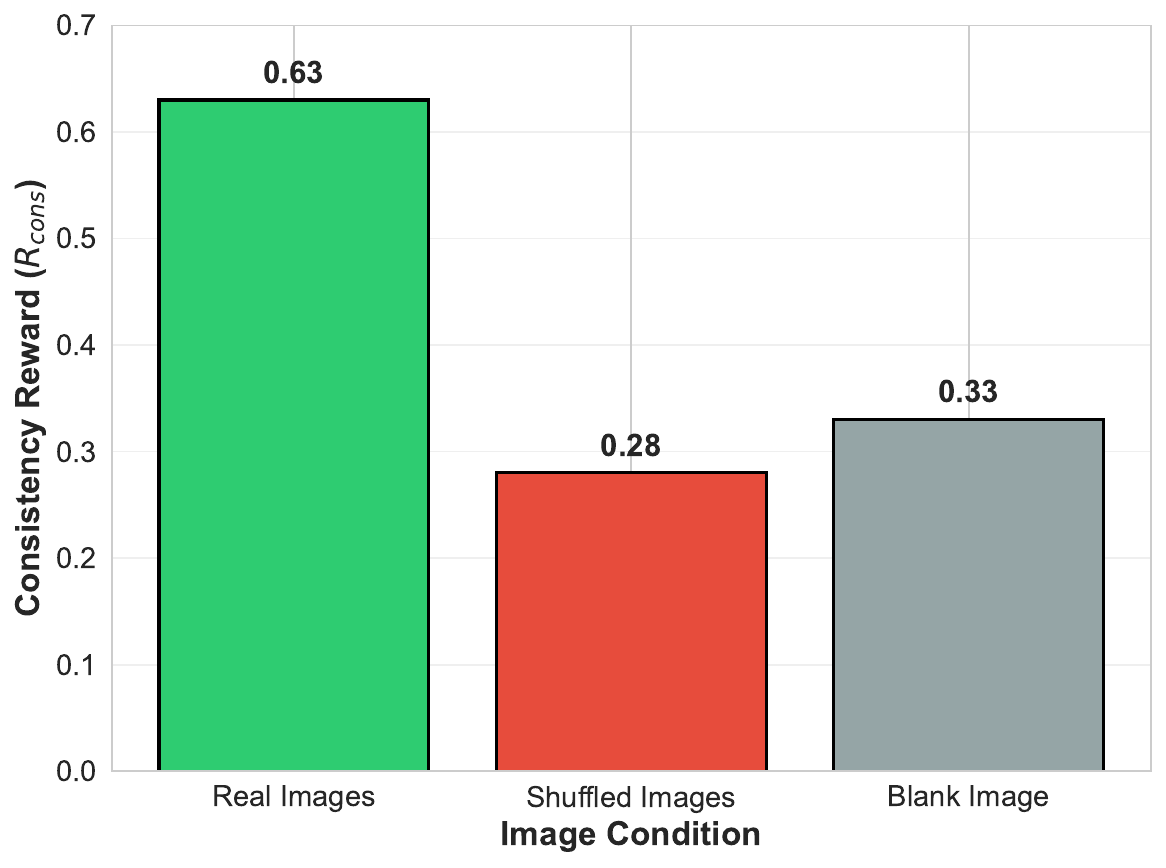}
        \caption{Consistency by image condition}
        \label{fig:consistency_conditions}
    \end{subfigure}
    \hfill
    \begin{subfigure}[b]{0.48\textwidth}
        \centering
        \includegraphics[width=\textwidth]{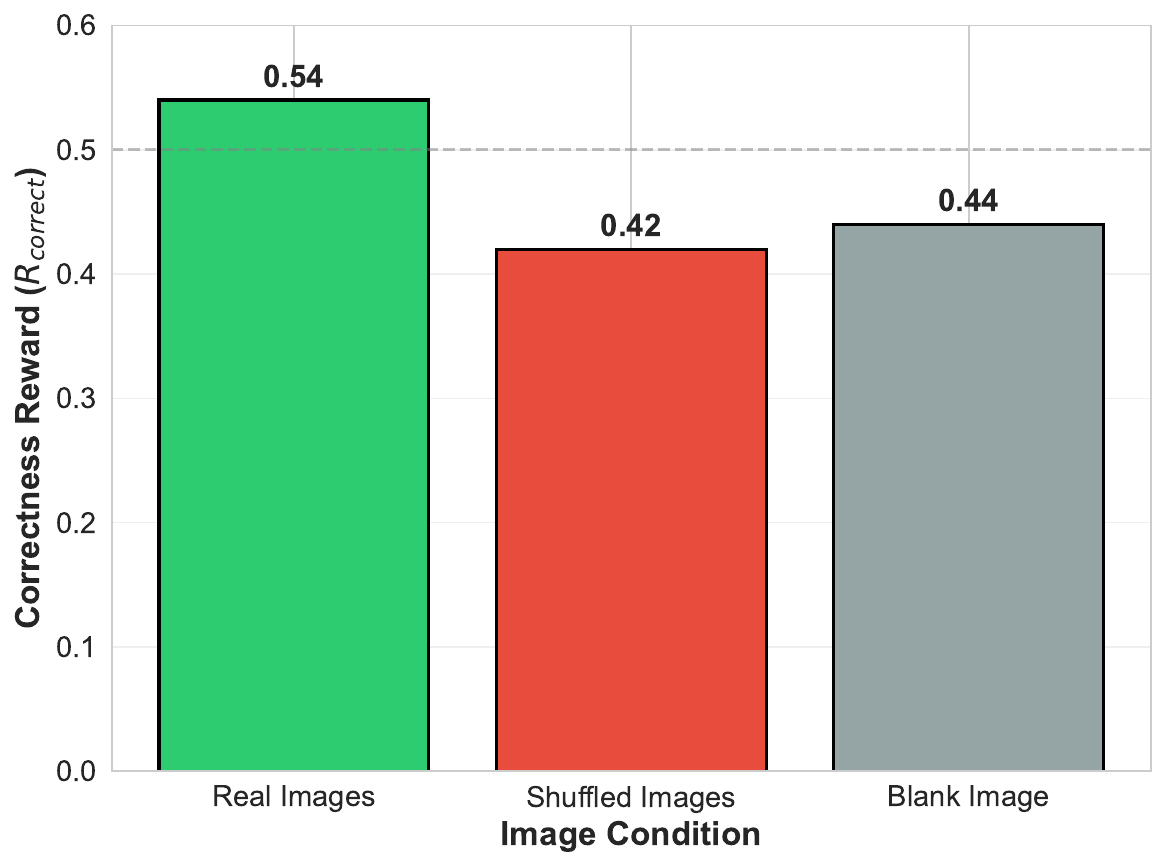}
        \caption{Correctness by image condition}
        \label{fig:correctness_conditions}
    \end{subfigure}
    \caption{\textbf{Consistency as a grounding signal.} (a) Mean $R_{\text{cons}}$ under three image conditions on an untrained base model. Shuffled images yield the lowest consistency, below even the no-image baseline. (b) Mean $R_{\text{correct}}$ shows degraded performance when visual grounding is disrupted, with both shuffled and blank conditions approaching random chance.}
    \label{fig:empirical_validation}
\end{figure}

Figure~\ref{fig:empirical_validation} validates our framework. Real images yield the highest consistency, shuffled images produce the lowest, and the no-image condition falls between them. The critical finding is the asymmetry: shuffled images perform worse than no images. With no image, the model hallucinates observations that may accidentally cohere with response claims. With a shuffled image, the model extracts observations that accurately describe the wrong scene, creating systematic conflict between observations and claims written for the correct image. The drop from real to shuffled demonstrates that $R_{\text{cons}}$ measures visual-reasoning alignment, not mere textual coherence. The correctness results (Figure~\ref{fig:correctness_conditions}) further show that consistency correlates with accuracy even without any consistency-based training. The parallel degradation in both metrics confirms that our grounding chain and consistency reward both incentivize visual-reasoning alignment.

\section{Main Result}
\label{sec:results}

We use Qwen3-VL-30B-A3B-Instruct \citep{qwen2025qwen3vl} as our base model. This is a mixture-of-experts model with 30B total parameters and 3B active parameters per token, providing a strong base with efficient inference. All training uses LoRA with rank 64. Cold-start SFT runs for 5 epochs with batch size 16 and learning rate $5 \times 10^{-5}$. GRPO training uses 32 candidate completions per prompt with temperature 0.7, max tokens 6,144, and the composite reward $R = R_{\text{format}} + R_{\text{correct}} + R_{\text{cons}}$. We train with batch size 16, learning rate $5 \times 10^{-5} \to 1 \times 10^{-7}$ cosine schedule.

We evaluate on MMRB2 \citep{hu2025mmrb2}, which consists of four subtasks, each containing 1,000 samples with binary preference labels: Text-to-Image (T2I) evaluates judgments of generated images against text prompts, Image Editing evaluates judgments of edit quality and instruction following, Interleaved Generation evaluates judgments of multi-turn conversations with images, and Multimodal Reasoning evaluates judgments requiring complex visual understanding and logical inference.

\begin{table}[http]
\centering
\caption{\textbf{Main results on MMRB2.} Accuracy (\%) across four subtasks. \mj achieves state-of-the-art with only 3B active parameters, surpassing all API-based and open-source models. Best results in \textbf{bold}, second-best \underline{underlined}.}
\label{tab:main_results}
\small
\begin{tabular}{@{}lccccc@{}}
\toprule
\textbf{Judge} & \textbf{T2I} & \textbf{Editing} & \textbf{Interleaved} & \textbf{Reasoning} & \textbf{Avg.} \\
\midrule
\multicolumn{6}{l}{\textit{Open-source multimodal LLMs}} \\
Gemma 3 4B \citep{gemma2025} & 51.7 & 51.0 & 51.3 & 48.8 & 50.7 \\
Gemma 3 12B \citep{gemma2025} & 56.0 & 58.0 & 58.0 & 49.3 & 55.3 \\
Gemma 3 27B \citep{gemma2025} & 58.3 & 60.2 & 61.1 & 49.4 & 57.3 \\
Qwen2.5-VL-7B \citep{bai2025qwen25vl} & 50.4 & 57.1 & 48.4 & 47.5 & 50.9 \\
Qwen2.5-VL-72B \citep{bai2025qwen25vl} & 59.1 & 64.6 & 62.3 & 50.0 & 59.0 \\
Qwen3-VL-8B \citep{qwen2025qwen3vl} & 59.4 & 61.7 & 61.5 & 54.6 & 59.3 \\
Qwen3-VL-32B \citep{qwen2025qwen3vl} & 64.1 & 67.3 & 70.5 & 56.6 & 64.6 \\
Qwen3-VL-30B-A3B \citep{qwen2025qwen3vl} & 60.0 & 59.5 & 57.3 & 57.3 & 58.5 \\
Qwen3-VL-235B-A22B \citep{qwen2025qwen3vl} & 62.0 & 64.8 & 69.0 & 55.9 & 62.9 \\
\midrule
\multicolumn{6}{l}{\textit{API-based Models}} \\
GPT-4o \citep{openai2024gpt4o} & 60.3 & 65.0 & 61.5 & 51.9 & 59.7 \\
GPT-4.1 \citep{openai2025gpt41} & 65.8 & 68.2 & 67.0 & 53.0 & 63.5 \\
GPT-5 \citep{openai2025gpt5} & 70.5 & 73.8 & 74.4 & 70.2 & 72.2 \\
Gemini 2.5 Flash \citep{google2025gemini25} & 63.1 & 66.5 & 69.4 & 57.5 & 64.1 \\
Gemini 2.5 Pro \citep{google2025gemini25} & 70.5 & 71.3 & \underline{75.1} & 66.6 & 70.9 \\
Gemini 3 Pro \citep{google2025gemini3} & \underline{74.4} & \underline{74.9} & \textbf{76.4} & \textbf{79.5} & \underline{76.3} \\
\midrule
\rowcolor{mj1green!12}
\textbf{MJ1} {\small (Qwen3-VL-30B-A3B + LoRA)} & \textbf{80.2} & \textbf{78.1} & 73.5 & \underline{76.4} & \textbf{77.0} \\
\bottomrule
\end{tabular}
\end{table}

Table~\ref{tab:main_results} shows results on MMRB2. \mj achieves 77.0\% overall accuracy, surpassing Gemini-3-Pro and all existing models. The gains are consistent across all four subtasks, demonstrating that our approach generalizes across the qualitatively different evaluation dimensions covered by MMRB2. With only 3B active parameters, \mj substantially outperforms models with orders of magnitude more parameters, reinforcing findings from J1 \citep{whitehouse2025j1} and JudgeLRM \citep{chen2025judgelrm} that the training recipe matters more than model scale for judgment tasks.

\section{Conclusion}
\label{sec:conclusion}

We present \mj, an RL-trained multimodal judge that achieves state-of-the-art performance on MMRB2 with only 3B active parameters. Our approach rests on two ideas: a grounded verification chain that front-loads visual observation extraction to mitigate attention decay, and a consistency reward that eliminates positional bias. We showed that structured grounding improves accuracy even without training, that our consistency mechanism incentivizes visual-reasoning alignment, and that the training recipe yields a model that surpasses orders-of-magnitude larger models at multimodal reward modeling.


\bibliographystyle{plainnat}
\bibliography{references}

@article{whitehouse2025j1,
  title={J1: Incentivizing Thinking in LLM-as-a-Judge via Reinforcement Learning},
  author={Whitehouse, Chenxi and Wang, Tianlu and Yu, Ping and Li, Xian and Weston, Jason and Kulikov, Ilia and Saha, Swarnadeep},
  journal={arXiv preprint arXiv:2505.10320},
  year={2025}
}

@article{hu2025mmrb2,
  title={Multimodal RewardBench 2: Benchmarking Reward Models for Omni-Modal Understanding and Generation},
  author={Hu, Yifan and others},
  journal={arXiv preprint arXiv:2512.16899},
  year={2025}
}

@article{shao2024deepseekmath,
  title={DeepSeekMath: Pushing the Limits of Mathematical Reasoning in Open Language Models},
  author={Shao, Zhihong and Wang, Peiyi and Zhu, Qihao and Xu, Runxin and Song, Junxiao and Bi, Xiao and Zhang, Haowei and Zhang, Mingchuan and Li, YK and Wu, Y and others},
  journal={arXiv preprint arXiv:2402.03300},
  year={2024}
}

@article{chen2025judgelrm,
  title={JudgeLRM: Large Reasoning Models as a Judge},
  author={Chen, Nuo and Hu, Zhiyuan and Zou, Qingyun and Wu, Jiaying and Wang, Qian and Hooi, Bryan and He, Bingsheng},
  journal={arXiv preprint arXiv:2504.00050},
  year={2025}
}

@inproceedings{saha2025evalplanner,
  title={Learning to Plan \& Reason for Evaluation with Thinking-LLM-as-a-Judge},
  author={Saha, Swarnadeep and Li, Xian and Ghazvininejad, Marjan and Weston, Jason E and Wang, Tianlu},
  booktitle={ICML},
  year={2025}
}

@article{yasunaga2025mmrb1,
  title={Multimodal RewardBench: Holistic Evaluation of Reward Models for Vision Language Models},
  author={Yasunaga, Michihiro and others},
  journal={arXiv preprint arXiv:2502.14191},
  year={2025}
}

@inproceedings{li2024vlrewardbench,
  title={VL-RewardBench: A Challenging Benchmark for Vision-Language Generative Reward Models},
  author={Li, Lei and others},
  booktitle={CVPR},
  year={2025}
}

@article{chen2024fastv,
  title={An Image is Worth 1/2 Tokens After Layer 2: Plug-and-Play Inference Acceleration for Large Vision-Language Models},
  author={Chen, Liang and others},
  journal={arXiv preprint arXiv:2403.06764},
  year={2024}
}

@inproceedings{zhang2024sparsevlm,
  title={SparseVLM: Visual Token Sparsification for Efficient Vision-Language Model Inference},
  author={Zhang, Yuan and others},
  booktitle={ICML},
  year={2025}
}

@inproceedings{zhang2025llavamini,
  title={LLaVA-Mini: Efficient Image and Video Large Multimodal Models with One Vision Token},
  author={Zhang, Shaolei and others},
  booktitle={ICLR},
  year={2025}
}

@inproceedings{fu2025hidden,
  title={Hidden in Plain Sight: Evaluating Abstract Shape Recognition in Vision-Language Models},
  author={Fu, Arjun and others},
  booktitle={COLM},
  year={2025}
}

@article{han2025selfconsistency,
  title={Do Vision-Language Models Really Understand Visual Language?},
  author={Han, Jiwan and others},
  journal={arXiv preprint arXiv:2410.00304},
  year={2025}
}

@inproceedings{huang2024opera,
  title={OPERA: Alleviating Hallucination in Multi-Modal Large Language Models via Over-Trust Penalty and Retrospection-Allocation},
  author={Huang, Qidong and others},
  booktitle={CVPR},
  year={2024}
}

@inproceedings{jiang2025devils,
  title={Devils in Middle Layers: Detecting and Mitigating Object Hallucination via Attention Analysis},
  author={Jiang, Zhangyu and others},
  booktitle={CVPR},
  year={2025}
}

@article{gemma2025,
  title={Gemma 3 Technical Report},
  author={{Gemma Team}},
  journal={arXiv preprint arXiv:2503.19786},
  year={2025}
}

@article{bai2025qwen25vl,
  title={Qwen2.5-VL Technical Report},
  author={Bai, Shuai and Chen, Keqin and Liu, Xuejing and Wang, Jialin and Ge, Wenbin and Song, Sibo and Dang, Kai and Wang, Peng and Wang, Shijie and Tang, Jun and others},
  journal={arXiv preprint arXiv:2502.13923},
  year={2025}
}

@article{qwen2025qwen3vl,
  title={Qwen3 Technical Report},
  author={{Qwen Team}},
  journal={arXiv preprint arXiv:2505.09388},
  year={2025}
}

@article{openai2024gpt4o,
  title={GPT-4o System Card},
  author={Hurst, Aaron and Lerer, Adam and Goucher, Adam P and Perelman, Adam and Ramesh, Aditya and Clark, Aidan and Ostrow, AJ and others},
  journal={arXiv preprint arXiv:2410.21276},
  year={2024}
}

@misc{openai2025gpt41,
  title={Introducing GPT-4.1 in the API},
  author={{OpenAI}},
  year={2025},
  howpublished={\url{https://openai.com/index/gpt-4-1/}}
}

@misc{openai2025gpt5,
  title={Introducing GPT-5},
  author={{OpenAI}},
  year={2025},
  howpublished={\url{https://openai.com/index/introducing-gpt-5/}}
}

@misc{google2025gemini25,
  title={Gemini 2.5: Our Most Intelligent AI Model},
  author={{Gemini Team}},
  year={2025},
  howpublished={\url{https://blog.google/technology/google-deepmind/gemini-model-thinking-updates-march-2025/}}
}

@misc{google2025gemini3,
  title={Gemini 3 Pro Technical Report},
  author={{Google DeepMind}},
  year={2025},
  howpublished={\url{https://deepmind.google/technologies/gemini/pro/}}
}

@misc{rapidata2024,
  title={Rapidata Coherence Dataset},
  author={{Rapidata}},
  year={2024},
  howpublished={\url{https://huggingface.co/datasets/Rapidata/Rapidata_Coherence}}
}

@misc{editreward2024,
  title={EditReward: A Preference Dataset for Image Editing},
  author={{EditReward Team}},
  year={2024},
  howpublished={\url{https://huggingface.co/datasets/EditReward}}
}

@article{rlaifv2024,
  title={RLAIF-V: Aligning MLLMs through Open-Source AI Feedback for Super GPT-4V Trustworthiness},
  author={Yu, Tianyu and others},
  journal={arXiv preprint arXiv:2405.17220},
  year={2024}
}

@article{ouyang2022training,
  title={Training language models to follow instructions with human feedback},
  author={Ouyang, Long and Wu, Jeffrey and Jiang, Xu and Almeida, Diogo and Wainwright, Carroll and Mishkin, Pamela and Zhang, Chong and Agarwal, Sandhini and Slama, Katarina and Ray, Alex and others},
  journal={NeurIPS},
  year={2022}
}

@article{zheng2023judging,
  title={Judging LLM-as-a-Judge with MT-Bench and Chatbot Arena},
  author={Zheng, Lianmin and Chiang, Wei-Lin and Sheng, Ying and Zhuang, Siyuan and Wu, Zhanghao and Zhuang, Yonghao and Lin, Zi and Li, Zhuohan and Li, Dacheng and Xing, Eric and others},
  journal={NeurIPS},
  year={2023}
}

@article{geirhos2020shortcut,
  title={Shortcut Learning in Deep Neural Networks},
  author={Geirhos, Robert and others},
  journal={Nature Machine Intelligence},
  volume={2},
  pages={665--673},
  year={2020}
}

@article{lanham2023measuring,
  title={Measuring Faithfulness in Chain-of-Thought Reasoning},
  author={Lanham, Tamera and others},
  journal={arXiv preprint arXiv:2307.13702},
  year={2023}
}


\newpage
\appendix
\section{Training Data}
\label{app:data}

We construct training data spanning domains aligned with MMRB2 evaluation categories (Table~\ref{tab:data}).

\begin{table}[http]
\centering
\caption{\textbf{Training data composition.} Samples, sources, and citations for each domain.}
\label{tab:data}
\small
\begin{tabular}{@{}llr@{}}
\toprule
\textbf{Domain} & \textbf{Source Datasets} & \textbf{Samples} \\
\midrule
Text-to-Image & Rapidata Coherence \citep{rapidata2024} & 12K \\
Image Editing & EditReward \citep{editreward2024} & 38K \\
Reasoning & RLAIF-V \citep{rlaifv2024} & 20K \\
\midrule
\textbf{Total} & & \textbf{70K} \\
\bottomrule
\end{tabular}
\end{table}

\section{Training Dynamics}
\label{app:training}

\begin{figure}[h!]
\centering
\includegraphics[width=0.55\textwidth]{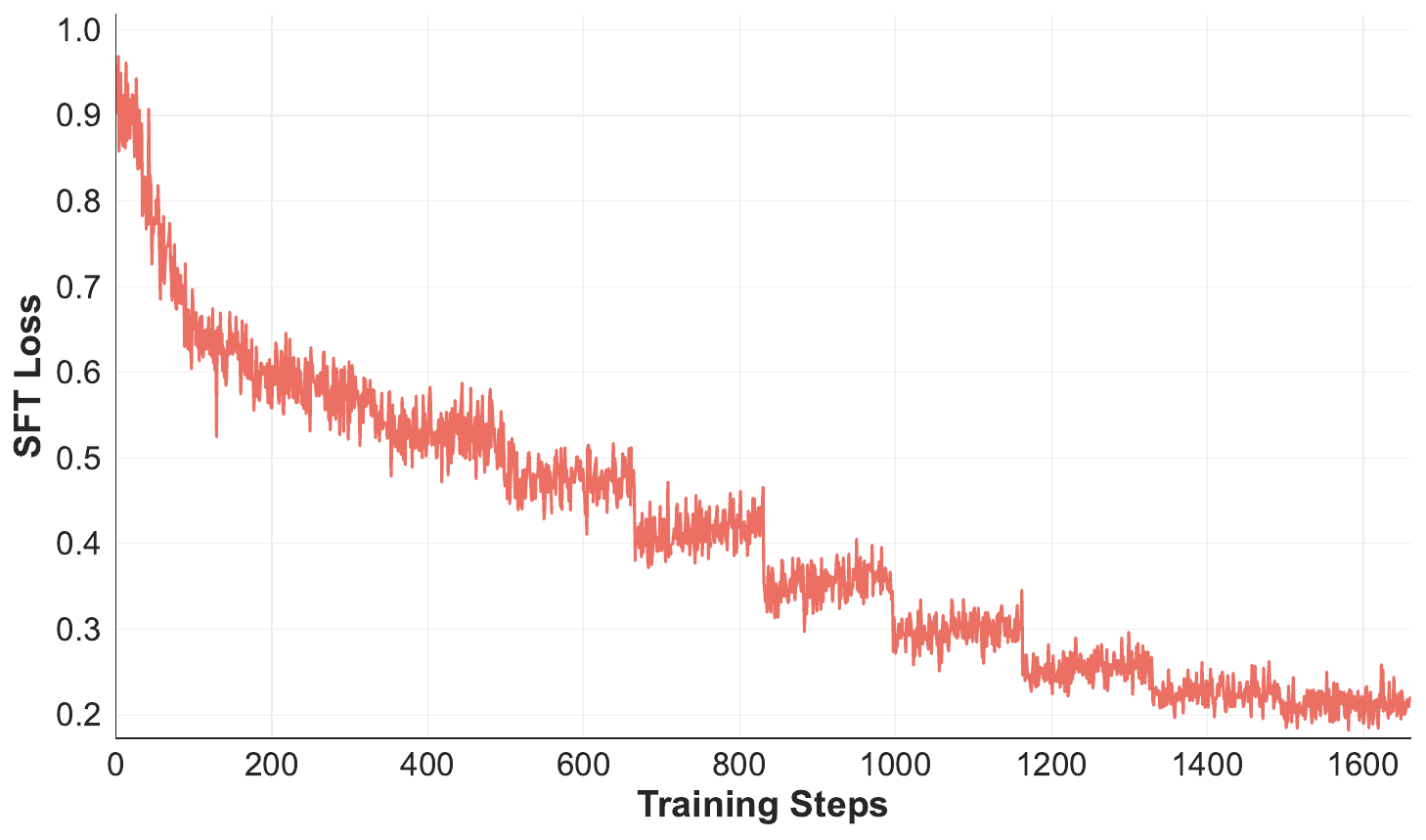}
\caption{\textbf{Cold-start SFT training loss.}}
\label{fig:sft_loss}
\end{figure}

\begin{figure}[h!]
    \centering
    \begin{subfigure}[b]{0.48\textwidth}
        \centering
        \includegraphics[width=\textwidth]{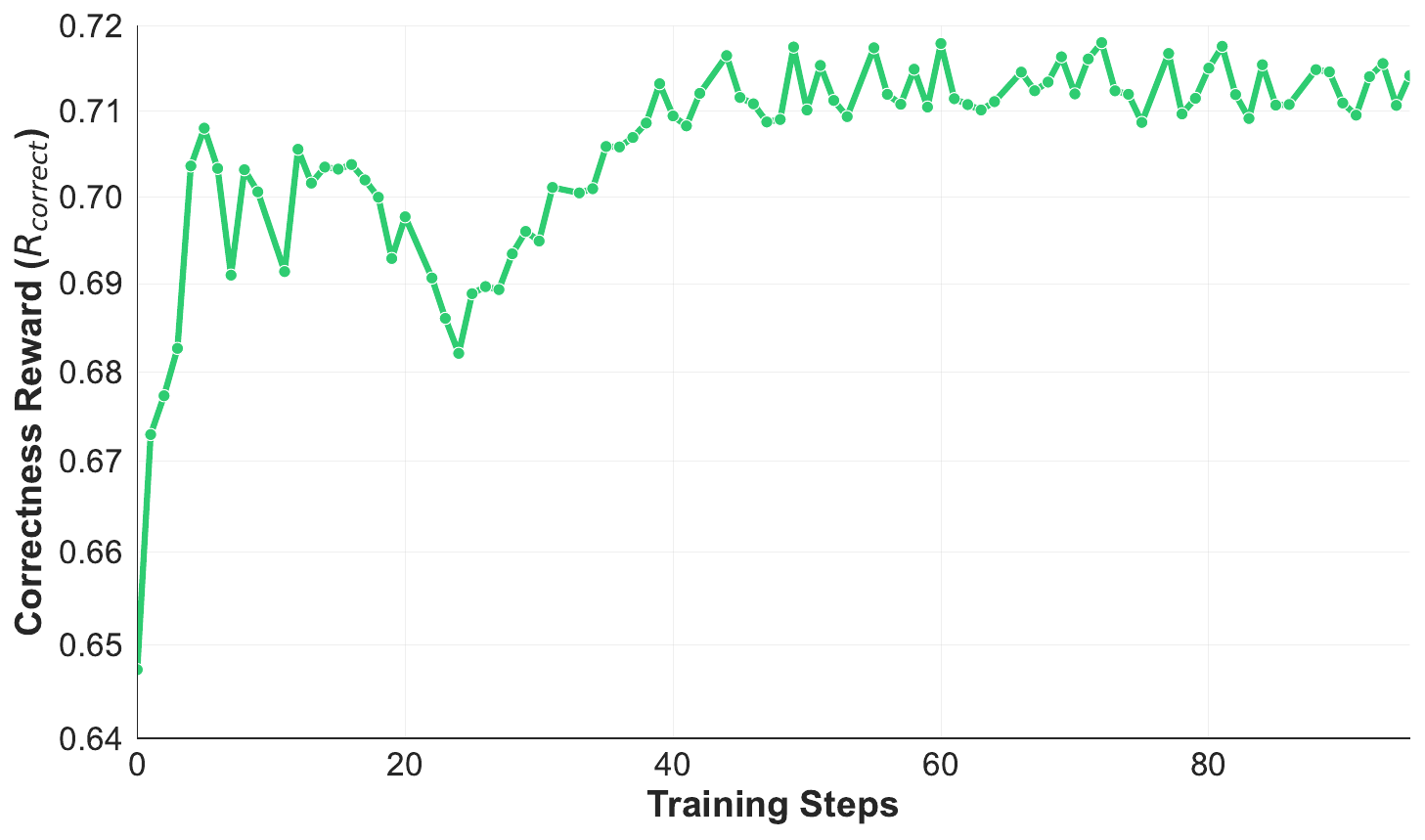}
        \caption{Correctness reward $R_{\text{correct}}$}
        \label{fig:reward_curve}
    \end{subfigure}
    \hfill
    \begin{subfigure}[b]{0.48\textwidth}
        \centering
        \includegraphics[width=\textwidth]{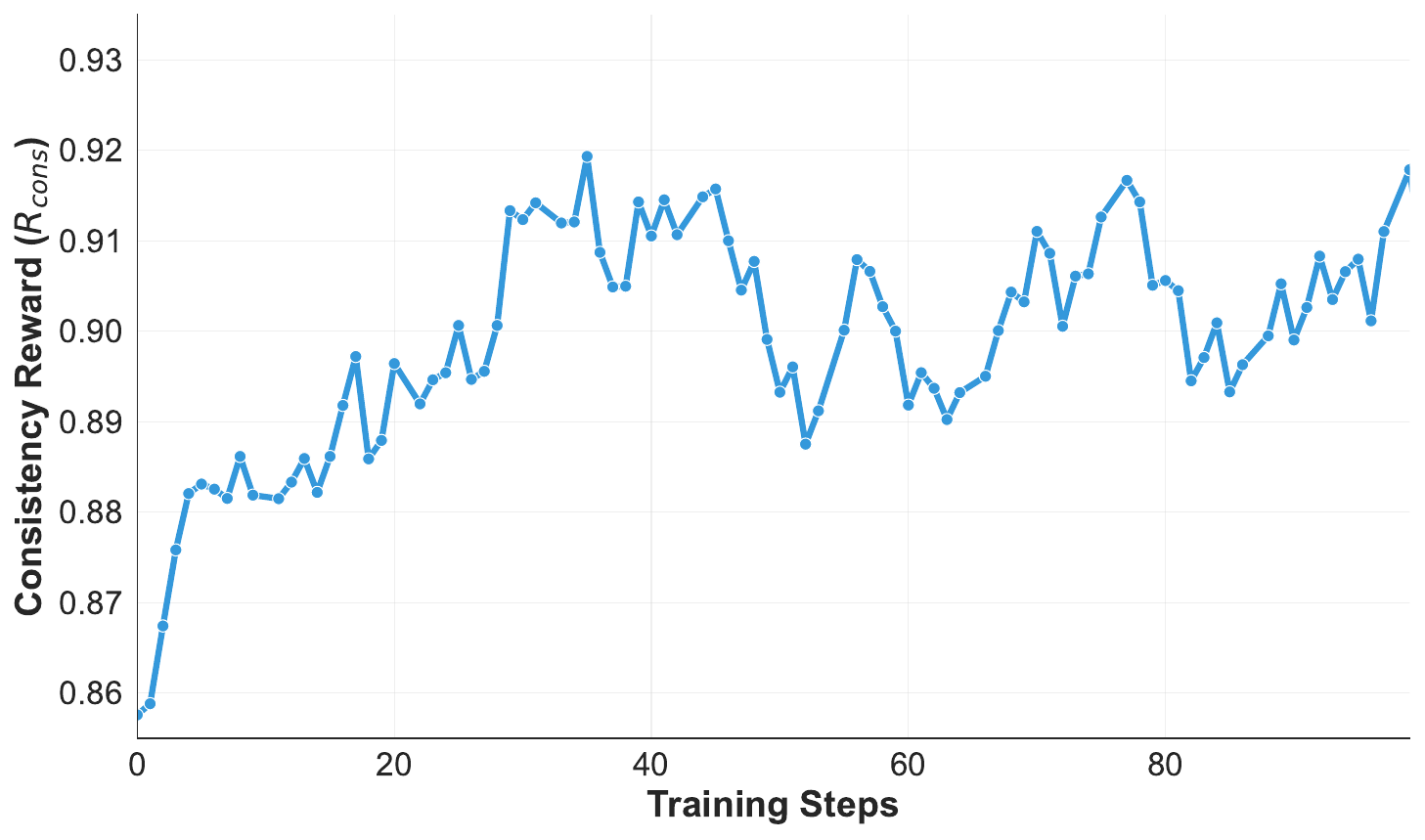}
        \caption{Consistency reward $R_{\text{cons}}$}
        \label{fig:consistency_reward}
    \end{subfigure}
    \caption{\textbf{GRPO reward curves.}}
    \label{fig:training_dynamics}
\end{figure}

Figure~\ref{fig:sft_loss} shows the smoothed cross-entropy loss during cold-start SFT on distilled reasoning traces. Figure \ref{fig:training_dynamics} shows the correctness and consistency rewards during the subsequent GRPO run.

\section{Prompt Templates}
\label{app:prompts}

\begin{tcolorbox}[
  title={\textbf{MJ1 Grounded Verification Prompt}},
  colback=backcolour,
  colframe=gray!60,
  fonttitle=\small\bfseries,
  breakable,
  left=4pt, right=4pt, top=4pt, bottom=4pt,
  boxrule=0.5pt
]
\begin{lstlisting}[style=xmlstyle, numbers=none, backgroundcolor=\color{backcolour}, xleftmargin=0pt, frame=none]
You are an expert in multimodal quality analysis and
generative AI evaluation. Your role is to act as an
objective judge for comparing two AI-generated responses
to the same prompt. You will evaluate which response is
better based on a comprehensive rubric.

**Reason through your evaluation using this structure:**
1. Extract key observations from any provided image(s).
2. Extract verifiable claims from each response.
3. Check whether each claim is consistent with your
   observations.
4. Using your consistency verification, evaluate both
   responses against each evaluation criterion to
   determine which performs better.
5. Provide final scores (no ties allowed).

**Evaluation Criteria:**
{EVALUATION_CRITERIA}

**REQUIRED OUTPUT FORMAT:**

<prompt_img_understanding>
[Describe what the prompt image shows (if it exists)]
</prompt_img_understanding>

<response_a_img_understanding>
[Describe what response A image shows (if it exists)]
</response_a_img_understanding>

<response_b_img_understanding>
[Describe what response B image shows (if it exists)]
</response_b_img_understanding>

<response_claims>
  <response_a_claims>
  [Verifiable claims from response A]
  </response_a_claims>

  <response_b_claims>
  [Verifiable claims from response B]
  </response_b_claims>
</response_claims>

<consistency_verification>
  <response_a_verification>
  [Verify Response A's claims against observations]
  </response_a_verification>

  <response_b_verification>
  [Verify Response B's claims against observations]
  </response_b_verification>
</consistency_verification>

<evaluate_criteria>
[For each criterion, evaluate both responses and explain
which performs better based on observations and
verification.]
</evaluate_criteria>

<scores>
\boxed{response_A_score, response_B_score}
</scores>

**Rules:**
- Scores are integers from 1 to 10 (higher is better)
- Scores must be different (no ties allowed)
- Higher accuracy and consistency = higher score
- Check for errors, hallucinations, and missing
  requirements
\end{lstlisting}
\end{tcolorbox}
\captionof{figure}{\textbf{MJ1 grounded verification prompt.} The five-stage structure (observations $\to$ claims $\to$ verification $\to$ evaluation $\to$ scores) enforces visual grounding by requirting the model to extract and cross-reference image content before scoring. The \texttt{\{EVALUATION\_CRITERIA\}} placeholder is filled with task-specific criteria at runtime.}
\label{fig:grounding_prompt}

\begin{figure}[h!]
\begin{tcolorbox}[
  title={\textbf{Text-to-Image Generation Criteria}},
  colback=backcolour,
  colframe=mj1blue!50,
  fonttitle=\small\bfseries,
  left=4pt, right=4pt, top=4pt, bottom=4pt,
  boxrule=0.5pt
]
\small
\begin{enumerate}[leftmargin=*, itemsep=1pt, parsep=0pt]
\item \textbf{faithfulness\_to\_prompt}: Which response better adheres to the composition, objects, attributes, and spatial relationships described in the text prompt?
\item \textbf{text\_rendering}: If either response contains rendered text, which one has better text quality (spelling, legibility, integration)? If none, state ``Not Applicable.''
\item \textbf{input\_faithfulness}: If an input image is provided, which response better respects and incorporates the key elements and style of that source image? If none, state ``Not Applicable.''
\item \textbf{image\_consistency}: If multiple images are generated, which response has better visual consistency between images? If none, state ``Not Applicable.''
\item \textbf{text\_image\_alignment}: Which response has better alignment between text descriptions and visual content?
\item \textbf{text\_quality}: If text was generated, which response has better linguistic quality (correctness, coherence, grammar, tone)?
\item \textbf{overall\_quality}: Which response has better general technical and aesthetic quality, realism, coherence, and fewer visual artifacts?
\end{enumerate}
\end{tcolorbox}
\caption{\textbf{Task-specific evaluation criteria for Text-to-Image Generation.}}
\label{fig:criteria_t2i}
\end{figure}

\begin{figure}[h!]
\begin{tcolorbox}[
  title={\textbf{Image Editing Criteria}},
  colback=backcolour,
  colframe=mj1green!50,
  fonttitle=\small\bfseries,
  left=4pt, right=4pt, top=4pt, bottom=4pt,
  boxrule=0.5pt
]
\small
\begin{enumerate}[leftmargin=*, itemsep=1pt, parsep=0pt]
\item \textbf{text\_faithfulness}: Which response better adheres to the text editing instruction? Consider how well each response follows the specific editing instructions (e.g., adding objects, changing colors, modifying scenes).
\item \textbf{image\_faithfulness}: Which response better preserves important aspects of the original image (composition, lighting, style, background elements) while making the requested changes?
\item \textbf{overall\_image\_quality}: Which response has better general technical and aesthetic quality, with fewer visual artifacts or inconsistencies introduced during editing?
\item \textbf{text\_rendering}: If either response contains rendered text, which one has better text quality (spelling, legibility, integration)? If none, state ``Not Applicable.''
\end{enumerate}
\end{tcolorbox}
\caption{\textbf{Task-specific evaluation criteria for Image Editing.}}
\label{fig:criteria_editing}
\end{figure}

\begin{figure}[h!]
\begin{tcolorbox}[
  title={\textbf{Interleaved Generation Criteria}},
  colback=backcolour,
  colframe=mj1purple!50,
  fonttitle=\small\bfseries,
  left=4pt, right=4pt, top=4pt, bottom=4pt,
  boxrule=0.5pt
]
\small
\begin{enumerate}[leftmargin=*, itemsep=1pt, parsep=0pt]
\item \textbf{text\_faithfulness}: Which response better adheres to the text instruction?
\item \textbf{image\_faithfulness}: Which response better respects and incorporates the key elements of the input image? If no input image, state ``Not Applicable.''
\item \textbf{overall\_image\_quality}: Which response has better overall quality of generated images?
\item \textbf{congruence}: If multiple images are generated, which response has better cross-generation visual consistency? If none, state ``Not Applicable.''
\item \textbf{text\_image\_alignment}: Which response has better generated text-image alignment?
\item \textbf{text\_quality}: If text was generated, which response has better linguistic quality? If none, state ``Not Applicable.''
\item \textbf{text\_rendering}: If either response contains rendered text within images, which has better correctness? If none, state ``Not Applicable.''
\end{enumerate}
\end{tcolorbox}
\caption{\textbf{Task-specific evaluation criteria for Interleaved Generation.}}
\label{fig:criteria_interleaved}
\end{figure}

\begin{figure}[h!]
\begin{tcolorbox}[
  title={\textbf{Visual Reasoning Criteria}},
  colback=backcolour,
  colframe=orange!50,
  fonttitle=\small\bfseries,
  left=4pt, right=4pt, top=4pt, bottom=4pt,
  boxrule=0.5pt
]
\small
\begin{enumerate}[leftmargin=*, itemsep=1pt, parsep=0pt]
\item \textbf{visual\_understanding}: Which response demonstrates better comprehension of the visual content? Consider accuracy in identifying objects, spatial relationships, colors, quantities, and other visual details.
\item \textbf{reasoning\_quality}: Which response shows stronger logical reasoning and analytical thinking? Evaluate the coherence of arguments, validity of inferences, and ability to connect visual observations to conclusions.
\item \textbf{accuracy}: Which response provides more accurate and factually correct information based on the provided images and context?
\item \textbf{completeness}: Which response more thoroughly addresses all aspects of the question?
\item \textbf{clarity}: Which response communicates its analysis more clearly and effectively?
\item \textbf{depth}: Which response provides deeper insight and analysis beyond surface-level observations?
\item \textbf{helpfulness}: Which response would be more helpful to someone trying to understand the visual content or solve the reasoning problem?
\end{enumerate}
\end{tcolorbox}
\caption{\textbf{Task-specific evaluation criteria for Visual Reasoning.}}
\label{fig:criteria_reasoning}
\end{figure}

\end{document}